# Classification dynamique d'un flux documentaire : une évaluation statique préalable de l'algorithme GERMEN.

Alain Lelu*, Pascal Cuxac**, Joel Johansson*

*LASELDI / Université de Franche-Comté

30 rue Mégevand – 25030 Besançon cedex

prénom.nom@univ-fcomte.fr

**INIST / CNRS

2 Allée du Parc de Brabois - CS 10310 - 54514 Vandoeuvre-lès-Nancy Cedex

cuxac@inist.fr

## Abstract

Data-stream clustering is an ever-expanding subdomain of knowledge extraction. Most of the past and present research effort aims at efficient scaling up for the huge data repositories. Our approach focuses on qualitative improvement, mainly for "weak signals" detection and precise tracking of topical evolutions in the framework of information watch – though scalability is intrinsically guaranteed in a possibly distributed implementation. Our GERMEN algorithm exhaustively picks up the whole set of density peaks of the data at time t, by identifying the local perturbations induced by the current document vector, such as changing cluster borders, or new/vanishing clusters. Optimality yields from the uniqueness 1) of the density landscape for any value of our zoom parameter, 2) of the cluster allocation operated by our border propagation rule. This results in a rigorous independence from the data presentation ranking or any initialization parameter. We present here as a first step the only assessment of a static view resulting from one year of the CNRS/INIST Pascal database in the field of geotechnics.

## Résumé.

L'extraction non supervisée et incrémentale de classes sur un flot de données (data-stream clustering) est un domaine en pleine expansion. La plupart des approches visent l'efficacité informatique. La nôtre, bien que se prêtant à un passage à l'échelle en mode distribué, relève d'une problématique qualitative, applicable en particulier au domaine de la veille informationnelle : faire apparaître les évolutions fines, les « signaux faibles », à partir des thématiques extraites d'un flot de documents. Notre méthode GERMEN localise de façon exhaustive les maxima du paysage de densité des données à l'instant t, en identifiant les perturbations locales du paysage à t-1 induites par le document présenté, et les modifications de frontières de classes en résultant. Son caractère optimal provient de son exhaustivité (à une valeur du paramètre de localité correspond un ensemble unique de maxima, et un découpage unique des classes par notre règle de propagation à partir des maxima) qui la rend indépendante de tout paramètre d'initialisation et de l'ordre d'arrivée des données. Nous évaluerons dans un premier temps cet algorithme sous son aspect statique, pour l'année 2003 du corpus documentaire « 10 ans de géotechnique dans la base Pascal » (CNRS/INIST).

**Mots-clés :** *data-stream clustering*, classification incrémentale.

# 1. Introduction

La classification automatique non supervisée (clustering) forme un domaine de recherche en soi, avec une longue histoire, et de très nombreuses méthodes constituant autant de variations autour de questions, parmi d'autres, telles que :

- quel(s)paramètre(s) : nombre de classes ? ou valeur d'un paramètre de finesse d'analyse ? Seuil(s) ?

- classification sur les lignes et/ou sur les colonnes d'un tableau individus × descripteurs ?

- classes strictes ? floues ? recouvrantes ? ou noyaux stricts + zones d'influence recouvrantes + *outliers* ?

- efficacité informatique ? passage possible à l'échelle des gisements et flux de données actuels ?

- robustesse, résistance au « bruit » ?

Pour rendre compte avec exactitude des évolutions temporelles, cruciales dans beaucoup de domaines d'application, en particulier ceux de la veille d'information, il est nécessaire à notre avis :

1) de partir d'une base stable, c'est-à-dire d'une classification :

- indépendante de l'ordre de présentation des données (exigence n°1),

- indépendante des conditions initiales, que ce soit d'un choix de « graines de classes » arbitraires ou dépendantes des données (exigence n°2),

- impliquant un minimum de paramètres, un seul si possible (paramètre de « zoom »), pour réduire l'espace des choix et tendre vers un maximum de vérifiabilité et de reproductibilité (exigence n°3).

2) d'ajouter aux contraintes d'une bonne classification celle de l'incrémentalité (exigence N°4), afin de saisir les évolutions au fil de l'eau : rectifications de frontières entre classes, apparition de nouvelles classes, voire de « signaux faibles »... Le caractère dynamique est intrinsèquement présent dans les analyses utilisant les liens de citation entre articles scientifiques (ou les liens hypertexte du Web). Pour qu'on puisse parler véritablement d'incrémentalité, il faut que le résultat de la classification soit indépendant de l'ordre des données présentées antérieurement (exigence N°5), tout en découlant des données antérieures par un historique pouvant faire l'objet d'interprétations.

Notre démarche a donc été de concevoir une méthode où la contrainte d'incrémentalité participerait à un tout cohérent, en vue d'aboutir à tout instant à une classification qui ait du sens, et dont la différence de représentation par rapport à l'instant précédent ne proviendrait que des effets du temps, et non du mélange de ceux-ci avec la variabilité propre de l'algorithme, ce qui n'est pas le cas avec les principales méthodes de classification non supervisée.

## 2. Etat de l'art

Nous passerons rapidement en revue les principales familles de méthodes de clustering, afin de les situer par rapport à nos exigences. Nous signalerons au passage certains de nos travaux plus anciens qui s'inscrivent dans ces cadres.

### 2.1. Méthodes hiérarchiques

Ces méthodes, divisives ou agglomératives, le plus souvent conviviales et efficaces, satisfont à nos exigences 1 à 3 d'unicité des résultats. Mais au regard de la qualité des partitions obtenues à un niveau donné de l'arbre, un consensus existe pour leur préférer les méthodes à centres mobiles (Lebart et al., 1982). Les déformations qu'impose un modèle hiérarchique de partitions emboîtées à une réalité ayant toutes chances de se rapprocher d'une organisation en treillis de partitions (par ex. treillis de Galois, pour des descripteurs binaires) expliquent sans doute ce constat.

### 2.2. Méthodes à centres mobiles

Les méthodes procédant par agrégation autour de centres mobiles, comme les K-means et leurs nombreuses variantes, font partie d'une famille basée sur l'optimisation d'un indicateur numérique global de qualité de la partition, dans laquelle prennent place les méthodes à mélanges de lois probabilistes explicites, utilisant la procédure EM (Expectation Maximization) – pour une revue, cf. (Buntine, 2002). Ce problème d'optimisation étant NP-difficile, on ne sait que les faire converger vers un optimum local qui dépend de leur initialisation (par ex. positions initiales des centres choisies arbitrairement, ou en fonction des données), voire de l'ordre des données. Ce qui les disqualifie vis-à-vis de notre exigence N°2 en théorie comme en pratique - notre méthode des K-means axiales, (Lelu, 1994), fait partie de cette famille - ; les optima locaux, à nombre de classes donné, révèlent à peu près les même classes principales souvent triviales, mais peuvent faire apparaître / disparaître / fusionner / éclater les classes d'effectifs moyens ou faibles souvent les plus intéressantes.

Un bon nombre de variantes incrémentales de ces méthodes ont été proposées, dont on trouvera une revue partielle dans (Gaber et al., 2005). Beaucoup sont issues de l'action DARPA « Topic Detection and Tracking », comme (Binztock, Gallinari, 2002), (Chen et al., 2003). Mais toutes se concentrent sur l'efficacité informatique pour traiter des flux de dizaines ou centaines de milliers de dépêches d'agences, ou d'enregistrements issus d'entrepôts de données ou d'Internet. C'est aussi dans ce cadre d'efficacité que Simovici et al. (2005) proposent un algorithme glouton pour optimiser un critère de qualité de partition original propre aux descriptions par variables nominales, qui évite d'introduire (ou de faire calculer de façon heuristique) le paramètre de nombre de classes propre à la plupart des méthodes de cette famille.

A signaler la proposition de (Saint Léger 1997) d'isoler les « descripteurs remarquables » caractérisant globalement une période de temps, donc les évolutions entre périodes (« termes émergents » et documents les contenant). L'auteur construit pour chaque période un indicateur de « pouvoir d'attraction » de chaque terme par agrégation de résultats obtenus par une classification des termes. Mais cette classification, basée sur plusieurs paramètres de seuil, ne présente pas de garantie d'optimalité ou reproductibilité pratique.

### 2.3. Méthodes basées sur la densité

A notre connaissance, seules les méthodes basées sur la densité satisfont à notre exigence d'unicité des résultats. Elles s'appuient sur la notion de densité d'un nuage de points, locale par définition puisque caractérisant le voisinage d'un point défini par un seuil de distance ou un nombre de plus proches voisins : étant donné 1) un nuage de points multidimensionnel, 2) une définition de la densité en chaque point de cet espace, 3) la valeur du paramètre de localité de cette fonction densité (son « rayon »), le paysage de densité qui en découle est unique et parfaitement défini. L'énumération de l'ensemble de ses pics – ou éventuels plateaux -représente un optimum absolu ; si on sait les repérer, ces pics balisent des noyaux homogènes de points ; dans les zones intermédiaires, on peut définir de diverses façons des zones d'influence de ces noyaux. (Trémolières 79, 94) a proposé un algorithme général, dit de percolation, indépendant de la définition de la densité et du type de données, pour délimiter rigoureusement les noyaux, les points-frontière ambivalents et les points atypiques. Il procède par baisse progressive du niveau de densité depuis le point le plus dense, et diffusion autour des noyaux qui apparaissent successivement. D'autres travaux retrouvent le même principe de repérage des noyaux denses, le plus souvent avec une définition spécifique de la densité, et d'extensions de diverses sortes à partir des noyaux : (Moody 2001), (Guénoche 2004), (Hader et Hamprecht 2003), (Batagelj 2002). A noter que ces méthodes peuvent se traduire en termes de partitionnement de graphe, car définir une densité implique d'avoir fixé des relations de voisinage, donc un graphe. DBSCAN, décrit dans (Ester et al. 1996) utilise une définition propre de la densité au moyen de deux paramètres, dont l'un fixe le seuil à partir duquel les noyaux sont constitués et étendus. Dans (Lelu 1994) nous décrivons une méthode de montées en gradient parallèles sur un paysage de densité (« inertie tronquée » = somme des carrés des projections tronquées des vecteurs-données normalisés sur les axes de classes) à formalisme neuronal, dénommée analyse en composantes locales.

Mais la pratique de cette méthode nous a montré qu'une densité définie par un seuil fixe de distance ou d'angle est aveugle à certaines classes avérées, mais peu denses, du fait des grandes disparités de densité rencontrées dans les applications réelles : nous avons alors recherché une mesure de densité adaptative, se coulant sur les différences de densité autour de la moyenne de la région considérée, rejoignant en cela les constats faits par d'autres auteurs, comme (Ertöz et al. 2003). Ils utilisent comme nous la notion de K plus proches voisins permettant de définir un « rayon adaptatif », sans préjuger de la définition de cette densité.

Ici aussi des méthodes incrémentales ont été proposées : cf. une version incrémentale de DBSCAN par (Ester et al. 1998), ou la méthode de (Gao et al. 2005), spécifique pour des descripteurs quantitatifs (âge, revenus, …) au sein desquels on peut découper et localiser des hyper-rectangles denses.

Mais c'est dans le domaine des protocoles auto-organisateurs de réseaux de communication radio dits « ad hoc » que le thème de l'incrémentalité pour le partitionnement dynamique de graphes évolutifs a été abordé de front - ex. : (Mitton, Fleury 2003) – avec des objectifs d'application assez différents des nôtres : les voisinages se modifient à chaque pas de temps, sans nécessairement comporter d'entrées ou sorties du réseau, l'optimalité est moins recherchée qu'une stabilité relative de la composition des classes (strictes) et de l'identité des chefs de classe (*clusterheads*). Le principe intéressant pour nous est celui d'un algorithme intégralement distribué, pour lequel la connaissance à l'instant t par chaque unité à classer de ses voisins et des voisins de ses voisins (2-voisinage) est suffisante pour déterminer à chaque instant l'existence et le découpage des classes.

## 3. Méthode proposée

### *3.1. Choix de transformation des données et de la nature de la densité*

Pour le choix de la densité, il est possible de s'inspirer du coefficient d'agrégation (*clustering coefficient*) bien connu de ceux qui étudient les grands graphes « de terrain », comme le Web, les réseaux informatiques ou d'interactions biologiques, et ceux de synonymie linguistique [cf. journées (GRM 2003)]. Il s'agit pour nous de la somme des similarités de tous les liens entre les n points du 1-voisinage, divisée par le nombre n.(n-1) de liens orientés possibles.

Cependant celle qui nous a donné empiriquement les meilleurs résultats est la simple somme des similarités de tous les liens orientés entre les n points du 1-voisinage, qui traduit la notion d'intensité absolue de liens de voisinage, en particulier de voisinage réciproque : ainsi un lien dans les deux sens entre points voisins compte double (la matrice d'incidence des K plus proches voisins n'est pas symétrique).

Pour le choix de la mesure des liens, nous avons opté pour le cosinus dans l'espace distributionnel [cf. (Lelu 2003)], lié à la distance de Hellinger [cf. (Domengès, Volle 1979), (Bacelar-Nicolau 2002)] ; les vecteurs-documents $\mathbf{x}_t$ sont normalisés comme suit :

$$\mathbf{x}_t : \{x_{it}\} \rightarrow \mathbf{y}_t : \left\{\sqrt{\frac{x_{it}}{x_{.t}}}\right\}$$

où $x_{.t}$ est la somme des composantes $x_{it}$ du vecteur $\mathbf{x}_t$

De bonnes propriétés en découlent, en particulier :

. équivalence distributionnelle (invariance de la représentation au regard de la fusion de descripteurs de mêmes profils relatifs),

. incrémentalité : la normalisation des vecteurs-documents évite que l'arrivée d'un nouveau document perturbe les similarités déjà calculées ; mais aussi le caractère très « creux » de ces vecteurs permet l'incrémentalité pour les descripteurs : le nombre de dimensions de l'espace des descripteurs peut croître au fur et à mesure que de nouveaux descripteurs (par ex. des termes) sont rencontrés dans les documents.

. extension aisée aux données négatives.

. les cosinus sont liés à la distance D de Hellinger : $\quad D(t,t')^2 = 2(1 - \cos(\mathbf{y}_t, \mathbf{y}_{t'}))$

### 3.2. Algorithme opérant sur un graphe orienté valué

Le partitionnement du graphe valué et orienté des K plus proches voisins peut être vu 1) comme le repérage de nœuds-germes principaux, ou « chefs de classe », nœuds localement plus denses que leur entourage, puis 2) comme l'extension de leur zone d'influence par rattachement unique ou partagé de leurs voisins de plus en plus éloignés. Nous avons publié et expérimenté sur des données documentaires [cf. (Lelu, François 2003)] un algorithme de percolation modifié, aboutissant à extraire 1) les points isolés, 2) les noyaux stricts exclusifs d'une classe, 3) les points ambivalents appartenant à plusieurs classes, tous ces points se projetant à des hauteurs diverses sur chaque axe de classe. Mais cet algorithme n'a pas la propriété d'incrémentalité.

Dans notre présente optique incrémentale, nous nous donnons l'état d'un graphe à l'instant t, dont les nœuds ont pu être caractérisés par leur densité, ainsi que par leur « couleur », c'est-à-dire par leur rattachement à un éventuel nœud « chef de classe », dont le numéro constitue l'étiquette. L'arrivée d'un nouveau nœud va perturber localement cet état : un certain nombre de nœuds dans le voisinage direct ou indirect du nouvel arrivant vont voir leur densité changer – ce changement du paysage de densité induisant à son tour un réajustement des zones d'influence des chefs de classe, voire des changements de chefs de classe, ou l'apparition de nouvelles classes. De façon imagée, des changements tectoniques locaux dans un massif montagneux peuvent redessiner des lignes de partage des eaux, voire créer ou fusionner des bassins versants.

De nombreuses règles d'héritage du numéro du (ou des) chef(s) de classe sont possibles. Mais dans tous les cas nous avons affaire à la mise à jour des « couleurs » d'un paysage de densité sous l'effet 1) de l'arrivée d'un nouveau point (changement structurel), 2) d'un changement localisé des densités (changement quantitatif).

Si la mise à jour de la couleur d'un point ne dépend que de la couleur de ses voisins « surplombants », de densité supérieure, alors à paysage de densité donné et à graphe de voisinage donné on n'obtiendra qu'un seul coloriage : l'attribution des classes sera lui aussi indépendant de l'ordre des données.

On peut rapprocher notre problème de plusieurs autres traités dans le domaine des algorithmes basés sur les graphes, comme les automates cellulaires, bien connus (l'état d'un nœud à l'instant t dépend de l'état de ses voisins à t-1), ou les graphes évoluants (les liens d'un nœud à l'instant t dépendent des liens de ses voisins à t-1), ces derniers utilisés dans le domaine de la robotique reconfigurable. Mais notre problème d'« automate de graphes » sur graphe évolutif ne semble pas avoir été abordé, à notre connaissance.

### 3.3. Algorithme incrémental GERMEN : modifications locales *des voisinages, des densités et des noyaux de classes.*

On construit progressivement et en la mettant constamment à jour une structure de donnée comportant pour chaque noeud la liste de ses voisins, sa densité, et son (ou ses) numéros de chef(s) de classe. A chaque arrivée d'un vecteur (nœud) nouveau, on calcule les changements de densité induits dans son 2-voisinage (il ne peut pas y en avoir ailleurs du fait de notre définition de la densité), puis les changements de chef(s) de classe induits. Pour ce faire, on met à jour et on parcourt itérativement la liste des nœuds susceptibles de changer de classe, compte tenu de la règle choisie pour l'extension des classes. Au pire cette liste peut comporter tous les documents antérieurs, mais elle ne peut que se vider (en pratique elle se stabilise autour de deux ou trois centaines d'éléments en moyenne quand le nombre d'éléments à classer dépasse un millier).

```
. Initialisation : le premier noeud de la séquence n'a aucun lien, a une densité nulle et
son propre numéro comme chef de classe.
LCC = Ø  // LCC est la liste des listes de chef(s) de classe pour chaque nœud //

. POUR chaque nouveau nœud :

        // changements de densité induits : //
        .calcul de ses 1- et 2-voisinages entrants et sortants, et des modifications de
        voisinages occasionnée par son arrivée ; d'où la liste LL des noeuds touchés par
        une création / suppr. /modif. de lien.
        .calcul de sa densité à partir de son 1-voisinage.
        .POUR tout noeud de LL, et tout noeud de son 1-voisinage :
        - nouveau calcul de sa densité
fin POUR

        // changements de chef(s) de classe induits : //
L = LL
TANT QUE  la liste L des noeuds susceptibles de changer d'état est non-vide :
        liste LS = Ø
        POUR chaque noeud de L trié par densité décroissante :
            ~ appliquer la règle de changement de classe en fonction des classes des
              voisins entrants surplombants (données par LCC) et de leurs densités.
            ~ si changement :
                .mettre à jour LCC pour le nœud courant
                .déterminer les surplombés éventuels (dont le noeud courant est surplombant)
                ; .incrémenter la liste LS des surplombés.
        fin POUR
        L=LS
fin TANT QUE
```

Exemples de règles :

A – tout noeud hérite du N° de chef de classe de son voisin entrant le plus surplombant (c.a.d. dont la densité dépasse la sienne propre et celle de tout son 1-voisinage) s'il y en a ; sinon, on crée une nouvelle classe.

B – tout noeud hérite du ou des N° de chef de classe de tous ses voisins surplombants, s'il y en a (possibilité d'appartenance à plusieurs classes) ; sinon, on crée une nouvelle classe.

Avec la règle A, tout noeud appartient à une seule classe, ou est isolé. Avec la règle B tout noeud peut appartenir à une seule classe (noyau), à plusieurs classes (nœuds ambivalents) ou être isolé. C'est cette dernière règle que nous utilisons dans nos applications documentaires, pour respecter au maximum la polysémie des mots et des textes.

Ces règles garantissent l'unicité du résultat, quel que soit l'ordre des nœuds auxquels on les applique, à la différence de l'algorithme de Trémolières où l'ordre est imposé par la baisse progressive du niveau de densité (les étiquettes de classe sont attribuées quand on « baisse le niveau de l'eau »).

*Nota 1* : la prise en compte des valeurs ex-aequo de similarité et densité est indispensable pour assurer l'indépendance par rapport à l'ordre des données. Les K plus proches voisins d'un noeud peuvent donc être en nombre supérieur à K…

De même, en cas de voisins de même densité et dominant leurs voisinages, nous convenons de leur attribuer le numéro du nœud le plus ancien.

*Nota 2* : une règle de bon sens pour éviter les liens entre documents sans rapport sémantique aucun, dans les premiers pas d'un passage incrémental ou pour des documents très atypiques, est de donner une limite à la prise en compte des K plus proches voisins : nous l'avons limitée aux documents de similarité supérieure au seuil de 0.1 ; prendre comme seuil la similarité moyenne, par exemple, empêcherait l'indépendance rigoureuse par rapport à l'ordre de présentation des données.

### *3.4. Illustration du contenu des classes par leurs descripteurs les plus saillants.*

Chaque lien établi par notre méthode est un produit scalaire entre vecteurs-documents normalisés, donc un cosinus, au sein duquel chaque descripteur d'occurrence non nulle dans l'un et l'autre des documents a sa part. D'où l'idée de mesurer l'importance d'un descripteur pour une classe en cumulant ses participations aux divers liens intra-classe seulement ; ce qui élimine les liens de la classe avec les autres classes, sources de non-spécificité (mais ces derniers liens peuvent être intéressants si on cherche à caractériser les similarités entre classes). Comme une classe, outre son noyau strict, peut comporter des documents communs avec d'autres classes, mais de densité inférieure à celles du noyau, nous pondérons cette participation par la moyenne géométrique des densités des extrémités du lien.

Plus formellement, si $x_{it}$ est le nombre d'occurrences du descripteur i dans le document N°t, de somme $x_{.t}$ pour l'ensemble de ces descripteurs, si $c_i(t,t')$ est la contribution pondérée du descripteur i au lien entre les documents t et t', de densités respectives $d(t)$ et $d(t')$, et $C_i(k)$ est la contribution relative du descripteur i à la classe k, dont on nomme *liens(k)* l'ensemble des liens internes, alors :

$$c_{i(t,t')} = \sqrt{d(t)d(t')} \cdot \sqrt{\frac{x_{it}}{x_{.t}}} \cdot \sqrt{\frac{x_{it'}}{x_{.t'}}} \qquad \text{et} \qquad C_i(k) = \frac{\sum_{liens(k)} c_{i(t,t')}}{\sum_k \sum_{liens(k)} c_{i(t,t')}}$$

On trouvera figure 1 ci-dessous un exemple d'affichage des descripteurs saillants d'une classe, par ordre de *Ci* décroissants.

---

*N° de la classe = N° d'ordre de son élément le plus dense (chef de classe)*
- *Contenu en documents* :
    Col. 1 : nombre de chefs de classe (1 pour les noyaux, n pour les multivalents)
    Col. 2 : N°de document
    Col. 3 : densité
    Col. 4 : titres des résumés
- *Illustration par les termes* :
    Col. 1 : participation pondérée à la somme des liens intra-classe (en pour mille)
    Col. 2 : terme

---

**Classe N° 89 :**

```
Noyaux :                                                    Termes illustratifs :
1  89  2.8854058  Manet.:."J'ai.fait.ce.que.j'ai.vu"......  144  peintre..........
1  63  2.433547   Renoir,."Il.faut.embellir"..............   93  peinture.........
1  24  1.7542332  Monet...................................   74  Paris............
1  44  1.4896605  Le.Douanier.Rousseau,.un.naïf.dans.la.ju   54  salon............
1   3  1.3989404  Seurat,.le.rêve.de.l'art-science........   41  monet............
1  72  1.3843564  Mirù.le.peintre.aux.étoiles.............   39  tableau..........
1  59  1.1211095  Pont-Aven,.l'Ecole.buissonnière.........   27  couleur..........
1  64  1.0549923  Vuillard,.le.temps.détourné.............   20  atelier..........
1  52   .9954685  Matisse,.une.splendeur.inouîe...........   20  ecole............
1  45   .6776769  Géricault,.l'invention.du.réel..........   20  portrait.........
1  25   .4496930  Toulouse-Lautrec,.les.lumières.de.la.nui   18  voyage...........
                                                            17  travail..........
Documents bivalents :                                       17  parisienne.......
2  65  1.0586247  Le.Corbusier,.l'architecture.pour.émouvo   16  amitié...........
2  73   .8939584  L'Invention.des.musées..................   15  rêve.............
2   4   .378152   Les.surréalistes,.une.génération.entre.l   15  guerre...........
2  12   .3246430  Jules.Vernes,.le.rêve.du.progrès........   13  lumière..........
                                                            10  toile............
Documents trivalents et plus :                              10  maître...........
3  79   .6364670  Einstein,.la.joie.de.la.pensée..........    9  renoir...........
5   7   .5430696  Les.feux.de.la.terre.:.Histoire.de.volca    9  théorie..........
```

*Figure 1 – Exemple de classe GERMEN (données : 193 résumés de la collection Gallimard Jeunesse, indexés par 888 termes) :*

### 3.5. Complexité informatique

Dans son implantation actuelle, chaque étape incrémentale coûte de l'ordre de O(n) en temps de calcul, donc l'algorithme complet qui les cumule est en O(n²). Des optimisations sont possibles dans le calcul des similarités et des K plus proches voisins, compte tenu du caractère très creux des vecteurs-données ; mais c'est surtout en l'exécutant en mode distribué, ce à quoi il est adapté par construction, qu'il pourrait être rendu très peu dépendant de la taille des données : on pourrait concevoir des serveurs Web maintenant à jour pour chaque page la liste de leurs descripteurs (liens ou mots), de leurs 2-voisines et de leur(s) chef(s) de classe(s), et organisant la communication entre elles et avec les pages extérieures… L'efficacité dépendrait alors surtout des protocoles de communication et du nombre moyen de descripteurs attribués à chaque page.

## 4. Spécificité du type de classification proposé, limites ergonomiques et remèdes possibles

Pour récapituler de façon plus concrète, notre méthode dégage, à partir d'un tableau [individus×descripteurs], plusieurs catégories d'éléments :

- des *noyaux*, ou points d'accumulation de la densité, de trois espèces :

  a) des regroupements de 2 à quelques dizaines d'individus très homogènes (« *noyaux multi-individus* »),

  b) des individus (« *nodules* ») seuls, mais chefs de classe d'éléments à appartenance multiple qui leur confèrent le caractère de « grumeau » de densité. A ce titre, ces concentra-tions locales de densité sur un ou plusieurs individus ont toutes les chances de constituer autant de germes auxquels viendront s'agréger les futurs individus amenés par le flux des vecteurs-données – d'où le nom GERMEN donné à notre méthode.

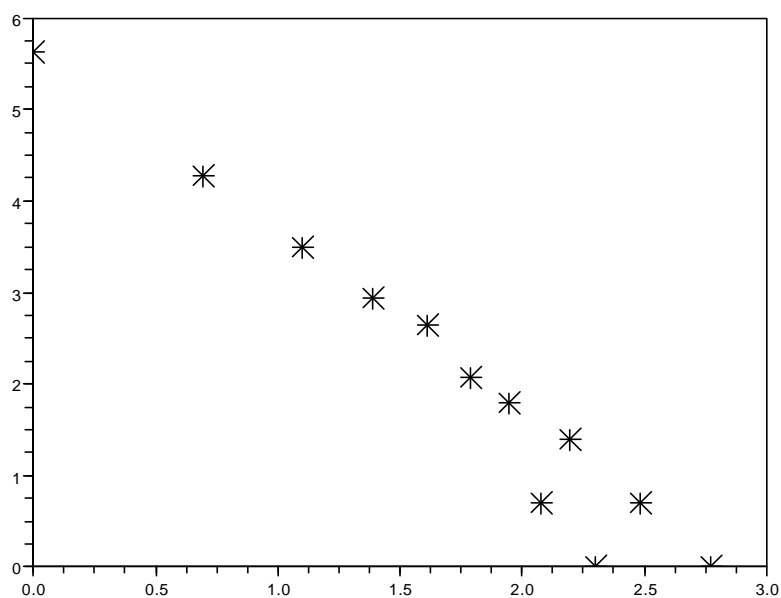

*Figure 2 : répartition des noyaux selon leur taille.
Abscisse : log(taille des noyaux), ordonnée : log(nombre des noyaux)*

c) des individus dits « *isolés* » n'ayant qu'eux-mêmes pour chefs de classe : à savoir tous les autres, de similitude nulle ou considérée comme négligeable par rapport aux noyaux extraits ; similitude insuffisante pour les inclure dans la liste des K plus proches voisins d'un élément d'appartenance stricte ou multiple à un noyau, donc empêchant de propager l'appartenance à ce noyau.

- des individus à appartenance multiple à deux ou plusieurs noyaux, dits « *multivalents* »,

Comme le confirmera expérimentalement la section 5 suivante, cette méthode conduit à un nombre important de noyaux multi-individus, mais surtout mono-individus (le graphique 2 plus haut en coordonnées log-log suggère une répartition en loi de puissance de la taille des noyaux – 280 noyaux de taille 1, 72 de taille 2,…).

Ceci montre que si notre méthode d'accrétion d'un flux de vecteurs-données revêt un caractère « naturel », comme le suggère cette répartition en loi de puissance - loi omniprésente dans les phénomènes sociaux et biologiques -, elle ne se conforme en aucune manière aux standards d'ergonomie d'une bonne classification attendus par l'esprit humain. En effet nul expert du domaine n'est prêt à travailler sur l'interprétation d'un nombre « déraisonnable » de classes – au-delà de quelques dizaines, une centaine au grand maximum – et si l'identification de quelques *outliers* est utile, leur excès est perturbant. D'autre part la notion de nodule à un seul élément est nouvelle, et même si elle s'apparente à celle de « signal faible » précurseur d'une future classe, leur nombre important empêche d'espérer les passer en revue utilement.

Nous avons conclu de tout cela qu'en parallèle avec le processus incrémental de formation et d'accrétion des noyaux, il est nécessaire de mettre en place un dispositif de « traduction » qui transforme toute vue instantanée prise avec GERMEN en une vue appréhendable par l'esprit humain, via le regroupement des noyaux, des individus multivalents et des isolés en un nombre raisonnable de classes, nombre éventuellement paramétrable directement ou indirectement, comme on va le voir ci-après.

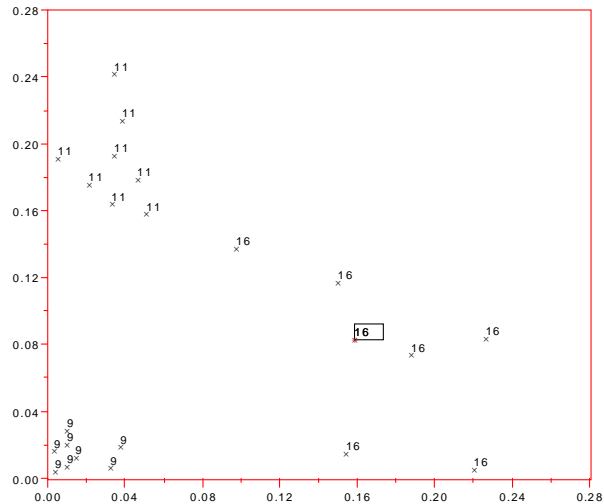

*Figure 3 – Exemple de classification par GERMEN dans deux dimensions sur trois sous-nuages de points de densités différentes. Les étiquettes de chaque point sont les numéros de leur chef de classe. Le dernier point traité est encadré.*

### *4.1. Traduction ergonomique d'une vue instantanée*

Notre pré-classification par extraction de noyaux et d'individus multivalents forme une base sur laquelle une ou plusieurs classifications plus agrégées et conformes aux attentes de l'esprit humain peuvent être construites. A ce jour nous n'avons testé qu'une seule idée, la plus immédiate : celle d'utiliser pour les regroupements les liens entre noyaux que constituent les individus multivalents. Nous avons développé le logiciel à interface interactive « Classotron » transformant une vue GERMEN à l'instant t en un ensemble de composantes connexes du graphe des liens entre noyaux, à valence fixée (par exemple, 2 pour les liens constitués par les individus bivalents). Ces composantes sont présentées à l'utilisateur, pour validation, par ordre décroissant de nombre d'individus concernés. Celui-ci peut visualiser le contenu des noyaux et des liens en individus et leur illustration par les descripteurs les plus saillants ; il a aussi la possibilité d'effectuer des regroupements manuels au sein de ces composantes connexes (cf. figure 4).

### *4.2. Discussion sur le paramètre K (nombre de plus proches voisins)*

Nous avons obtenus nos meilleurs résultats empiriques avec K=3, quelle que soit la taille des données et la dimension de l'espace de leurs descripteurs : nuage de points dans 2D, ou 193 résumés $\times$ 888 mots, ou 1175 documents $\times$ 3731 mots.

Diminuer ce nombre accentue la fragmentation de nombreux noyaux. L'augmenter tend à concentrer beaucoup d'individus sur quelques grosses classes peu homogènes, par un effet de chaîne marqué, sans diminuer sensiblement le nombre de nodules et d'isolés.

## 5. Application à la base documentaire « 10 ans de géotechnique dans la base Pascal »

Pour illustrer le type de classification opérée, on trouvera d'abord ci-dessus (figure 3) un exemple-jouet de classification de 3 sous-nuages de points de densités différentes dans 2 dimensions (plus une dimension « biais » de valeur constante et égale à 20) ; ici K=3 plus pro-ches voisins.

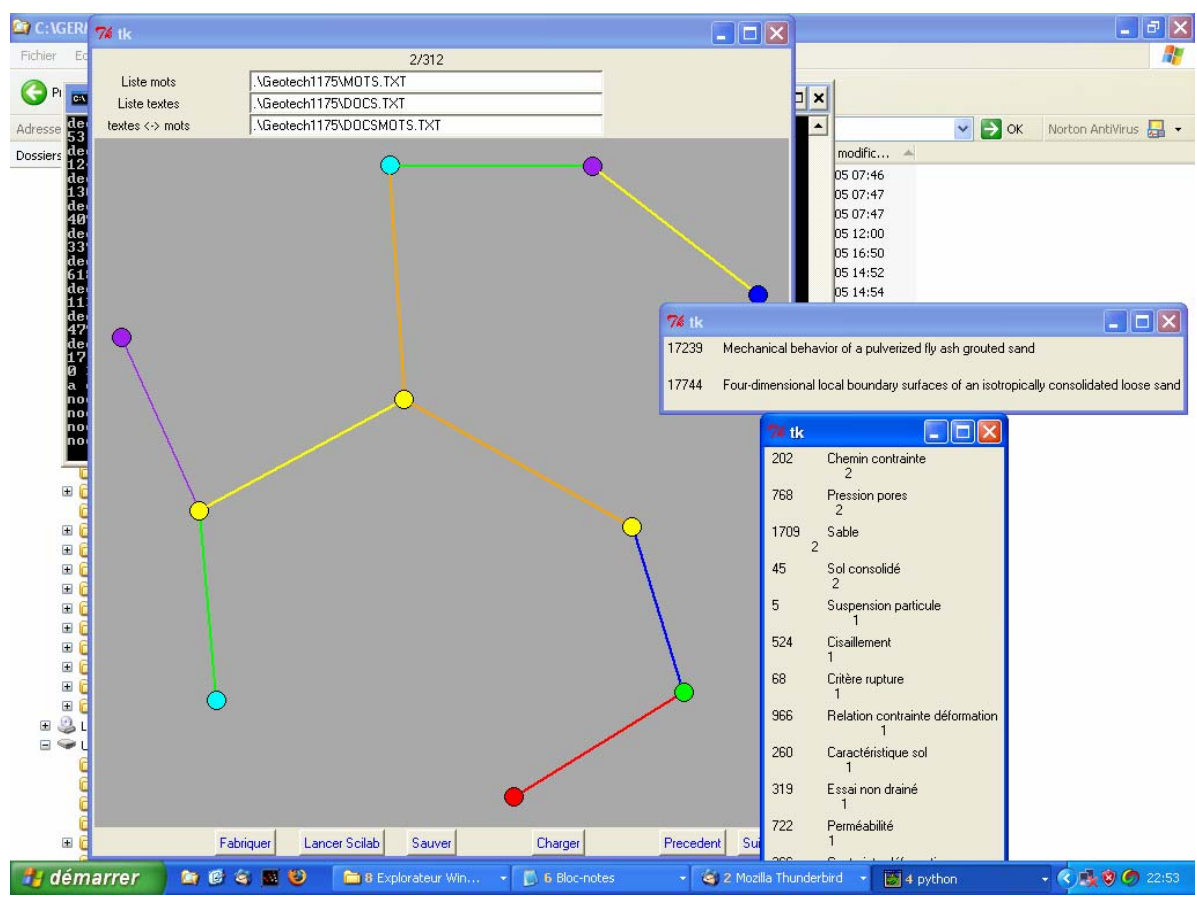

*Figure 4 – Ecran Classotron de base, avec un exemple de composante connexe à 10 noyaux, et un exemple de contenu de noyau, en documents et en mots.*

Les numéros sont ceux des chefs de classe de chaque point ; le dernier point pris en compte est encadré.

Nous avons constitué une base de test à partir d'un extrait d'une quinzaine de milliers d'enregistrements de la base Pascal de l'INIST (www.inist.fr) sur le thème de la géotechnique, de 1994 à 2004. Avant de travailler sur les aspects dynamiques, nous avons voulu nous assurer que les « photos » étaient bonnes ; pour cela nous avons entrepris l'évaluation statique de notre méthode sur 1175 documents de l'année 2003, décrits par 3731 termes d'indexation manuelle, une fois éliminés les hapax, les termes génériques et hors sujet.

Comme précisé ci-dessus, nous avons fixé à 3 le paramètre K. Nous avons obtenu 162 noyaux multidocuments, rassemblant 581 documents (49% du total) et de taille 2 à 16, 90 nodules unidocuments, 190 documents isolés (16%), et 314 documents multivalents (27%). Le graphique de la figure 2 donne la répartition de taille de tous les noyaux, nodules et isolés.

Notre interface Classotron (cf. figure 4) a permis d'obtenir une présentation « prédigérée », raisonnée de ces noyaux pour l'un d'entre nous, expert du domaine géotechnique, ce qui lui a permis 1) de valider l'ensemble des noyaux obtenus, 2) de valider ou invalider les regroupements par composantes connexes établis à partir des documents bivalents (nœuds = noyaux, liens = documents bivalents). Il a pu ainsi en un temps raisonnable établir une classification « idéale » sur les plus saillantes de ces données, en éliminant d'emblée le « marais » des documents trop isolés ou trop ambivalents. 46 classes ont été dégagées à partir de nos 54 composantes connexes de plus de 2 documents, chacune dotée d'un libellé synthétisant son contenu, comme « Cisaillement des roches fracturées », « Stabilité des versants », « Stabilité des remblais : étude des déformations », …

Une fois cette classification établie, il a été possible de la confronter à notre regroupement des noyaux en composantes connexes – mais la comparaison à tout autre regroupement obtenu par d'autres méthodes sera toujours possible.

Une méthode d'évaluation standard en documentation automatique et apprentissage supervisé consiste à calculer les indicateurs « rappel » (dit aussi « couverture ») et « précision » :

Rappel = nombre d'éléments corrects trouvés / nombre d'éléments corrects

Précision = nombre d'éléments corrects trouvés / nombre d'éléments trouvés

Nous avons calculé ces taux de rappel et précision pour toutes les classes idéales définies, rangées par ordre de précision décroissante. La figure 5 rassemble ces résultats, et montre que GERMEN classe avec une excellente précision un sous-ensemble de 300 documents, moins dispersés que les autres. Mais la procédure d'agrégation des noyaux par les documents bivalents créée un agrégat dominant de 67 noyaux qui créée un « trou » considérable dans le rappel. Il faudra donc tester d'autres procédures d'agrégation.

## 6. Conclusion et perspectives

Nous avons présenté un algorithme de classification non supervisée répondant aux exigences d'un suivi dynamique rigoureux d'un flux de documents : il est à la fois optimal, au sens de la description exhaustive d'un paysage de densité adaptative, et incrémental. Son efficacité actuelle est suffisante pour traiter un flux de plusieurs milliers de documents, et une évaluation est présentée ici sur une base documentaire de cet ordre de grandeur, avec une interface utilisateur spécifique. Son passage à l'échelle est possible en mode distribué, par construction. Nous avons pris le parti d'un algorithme strictement déterministe dont l'avantage est l'adaptation parfaite à toute loi et changement de loi de répartition des données ; mais l'inconvénient de cet avantage est la variabilité des représentations instantanées, le « bruit » engendré par une fidélité trop grande à des données quand leur caractère est parfois largement arbitraire (par exemple un mot plutôt qu'un autre, dans le domaine de l'analyse textuelle ou dans l'indexation documentaire manuelle…). Dans le futur, il sera sans doute souhaitable d'introduire une part de « flou » au sein du processus, en veillant à ne pas retomber dans les problèmes d'optima locaux propres aux méthodes à modèles statistiques explicites.

Le processus pourra aussi être rendu décrémental, sur le même principe (suppression d'un point du nuage avec les modifications locales des liens, des densités, et des chefs de classe qui en découlent), fonction utile pour certaines applications (fenêtre glissante temporelle, « oubli » de données, traitement de flux de vecteurs entrants et sortants, …), ou sensible aux seules créations / modifications / suppressions de liens, à nombre de noeuds constant (par ex., pages Web décrites par leurs liens).

L'algorithme peut aussi s'appliquer directement à toute structure de graphe orienté – valué ou non - évolutive (liens bibliométriques de citation, liens Web, …).

Nous avons vu que le type de classification obtenue ne se prête pas directement à l'interprétation humaine : il reste à améliorer le processus de traduction de celle-ci en un nombre raisonnable de classes, processus qui nous a permis de constituer au passage une base de test utile à tous ceux qui ceux qui se préoccupent d'évaluation de méthodes de classification non supervisées. Une fois complétée par l'examen des nodules et individus isolés, cette base de test sera mise à disposition publique sur le site Web de l'INIST <http://visa.inist.fr/~cuxac/JADT>.

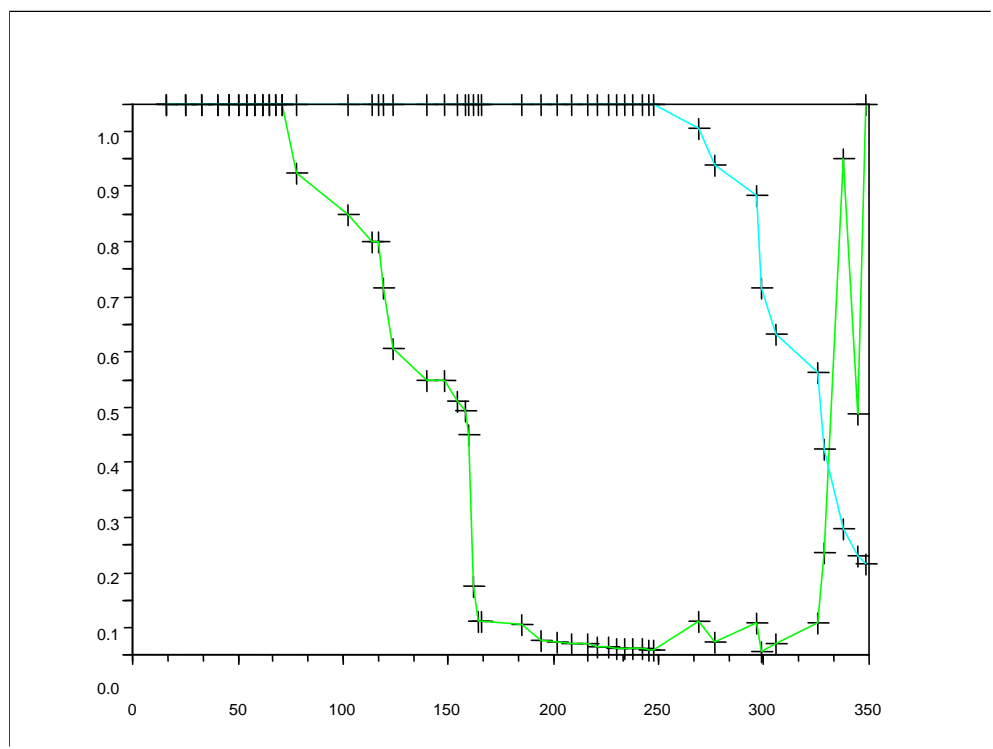

*Figure 5 : Evaluation de la classification statique opérée sur l'année 2003*
*En abscisse : nombre de docs, classés par précision décroissante des noyaux auxquels ils*
*appartiennent. En vert : rappel ; en bleu : précision*

Mais il reste surtout à explorer en grandeur réelle les diverses applications possibles, avec les problèmes qui vont avec : efficacité informatique, ergonomie de représentation en fonction des attentes des usagers, et définition même de ce que peut être la représentation dynamique et interactive d'une réalité évolutive - problème peu abordé jusqu'à présent puisqu'une telle représentation est impossible sur une feuille de papier, qu'elle est limitée au défilement séquentiel avant et arrière sur support audiovisuel, et que peu de pistes ont été explorées en ce qui concerne l'écran d'ordinateur.

## Références